\documentclass[letterpaper]{article} 
\usepackage{aaai2026}
\usepackage{times}  
\usepackage{helvet}  
\usepackage{courier}  
\usepackage[hyphens]{url}  
\usepackage{booktabs}
\usepackage{multirow}
\usepackage{xcolor}
\usepackage{colortbl}
\usepackage{enumitem}
\usepackage{amssymb}
\usepackage{pifont}
\newcommand{\xmark}{\ding{55}}  
\newcommand{\cmark}{\ding{51}}  
\usepackage{tabularx} 
\usepackage[most]{tcolorbox}
\usepackage{xcolor}

\definecolor{boxcolor}{RGB}{36,64,139} 
\usepackage{graphicx} 
\usepackage{natbib}  
\usepackage{caption} 
\usepackage{algorithm}
\usepackage{algorithmic}
\usepackage{amsmath}

\usepackage{newfloat}
\usepackage{listings}
\DeclareCaptionStyle{ruled}{labelfont=normalfont,labelsep=colon,strut=off} 
\floatstyle{ruled}
\newfloat{listing}{tb}{lst}{}
\floatname{listing}{Listing}
\title{VideoSeg-R1:Reasoning Video Object Segmentation via Reinforcement Learning}

\author{
    Zishan Xu\textsuperscript{\rm 1,*},
    Yifu Guo\textsuperscript{\rm 2,*},
    Yuquan Lu\textsuperscript{\rm 2,*},
    Fengyu Yang\textsuperscript{\rm 3},
    Junxin Li\textsuperscript{\rm 2}
}

\affiliations{
   \textsuperscript{\rm 1}Shanghai Jiao Tong University\\
       \textsuperscript{\rm 2}Sun Yat-sen University\\ \textsuperscript{\rm 3}South China Normal University\\
}

\usepackage{bibentry}
\usepackage[capitalize]{cleveref}
\crefname{section}{Section}{Sections}
\Crefname{section}{Section}{Sections}
\crefname{subsection}{Sec.}{Secs.}
\Crefname{subsection}{Section}{Sections}
\crefname{figure}{Figure}{Figsures}
\Crefname{figure}{Figure}{Figures}
\crefname{table}{Table}{Tables}
\Crefname{table}{Table}{Tables}

\begin{document}

\maketitle

\begin{abstract}
Traditional video reasoning segmentation methods rely on supervised fine-tuning, which limits generalization to out-of-distribution scenarios and lacks explicit reasoning. To address this, we propose \textbf{VideoSeg-R1}, the first framework to introduce reinforcement learning into video reasoning segmentation. It adopts a decoupled architecture that formulates the task as joint referring image segmentation and video mask propagation. It comprises three stages: (1) A hierarchical text-guided frame sampler to emulate human attention; (2) A reasoning model that produces spatial cues along with explicit reasoning chains; and (3) A segmentation-propagation stage using SAM2 and XMem. A task difficulty-aware mechanism adaptively controls reasoning length for better efficiency and accuracy. Extensive evaluations on multiple benchmarks demonstrate that VideoSeg-R1 achieves state-of-the-art performance in complex video reasoning and segmentation tasks. The code will be publicly available at \url{https://github.com/euyis1019/VideoSeg-R1}.
\end{abstract}
\section{Introduction}\label{sec:intro}
\textbf{Referring video object segmentation (RVOS)} requires a model to
\emph{localize} and \emph{segment} \textbf{one or multiple target objects}
throughout an entire video, given a natural-language description~\cite{guo2019video}.
Success hinges on two intertwined capabilities:
(i)~fine-grained \textit{spatial} precision at the pixel level and
(ii)~robust \textit{temporal} reasoning to track objects under motion,
occlusion, and appearance change.
In recent years, large language models (LLMs) have made remarkable progress across various dimensions~\cite{luo2025codetestcasesenough,lin2025seagentselfevolutiontrajectoryoptimization,du2025mokgr,du2025graphoracle,du2025graphmaster}, which in turn has driven the advancement of multimodal models. While recent multimodal large language models (MLLMs) excel on
static-image tasks, they falter when confronted with
long-form videos and complex language queries that demand multi-step
reasoning~\cite{cai2024survey}.

Existing RVOS pipelines overwhelmingly rely on
\textbf{supervised fine-tuning (SFT)}~\cite{wu2022language,lin2024villa}.
Although effective on in-distribution data,
SFT models (i)~overfit to seen categories and viewpoints~\cite{zhang2025parameter},
(ii)~lack interpretable reasoning chains.
Consequently, accuracy plummets when a query involves subtle temporal
context (e.g., “\emph{the man who appears after the car turns left}”) or
commonsense inference~\cite{bellver2022closer}.

\textbf{Reinforcement learning (RL)} has recently emerged as a powerful
tool for endowing language models with reasoning abilities.
Algorithms such as Group Relative Policy Optimization (GRPO)~\cite{GRPO}
have advanced chain-of-thought generation,
and their multimodal extensions have begun to tackle pixel-level vision
tasks.
\emph{Yet no prior work} has transferred RL-based reasoning to the
video reasoning segmentation domain, where the action space explodes with temporal length and
the reward must balance spatial accuracy against temporal consistency.

\begin{figure}[t]
    \centering
    \includegraphics[width=0.48\textwidth]{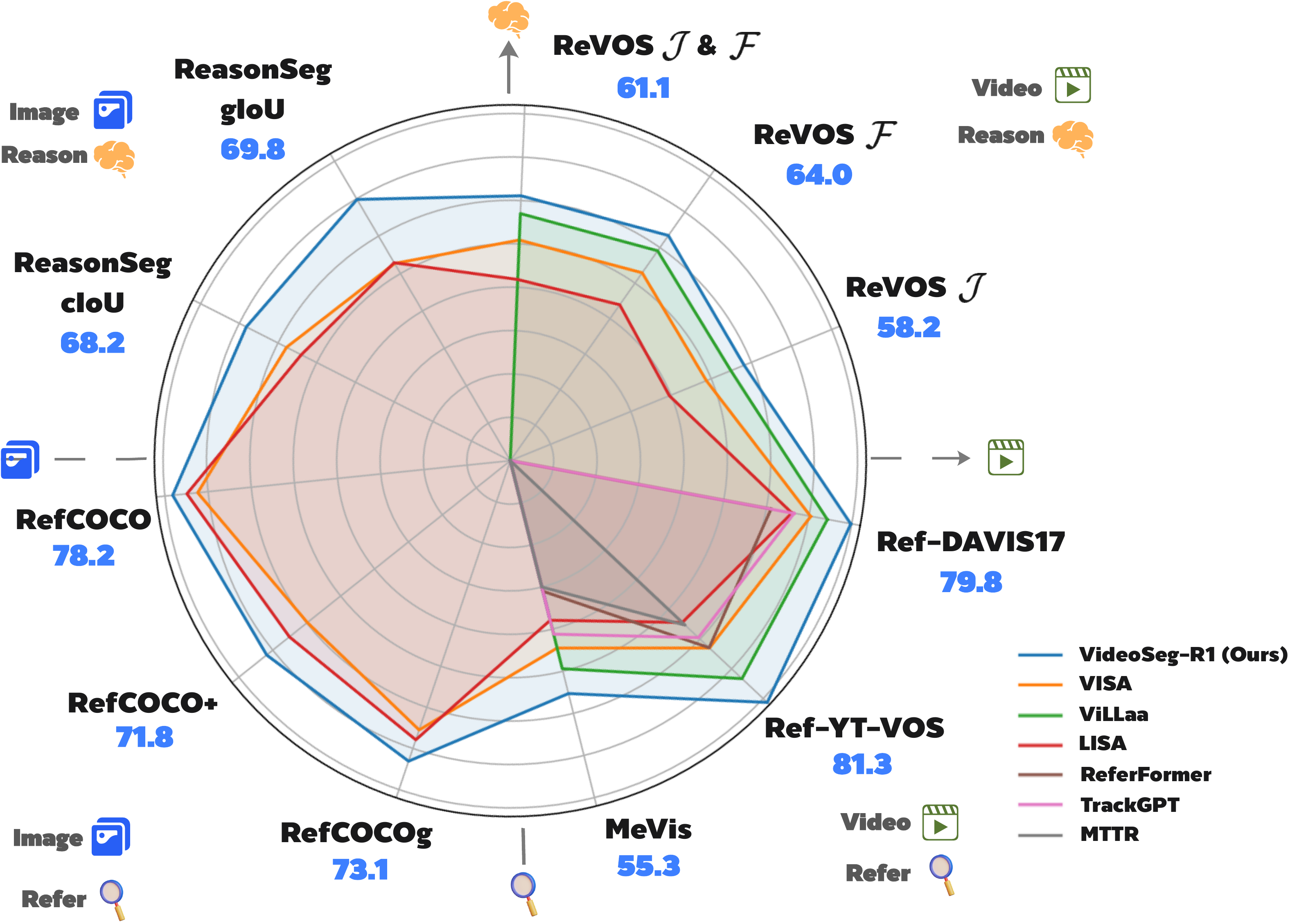}
    \caption{VideoSeg-R1 achieves state-of-the-art performance on both video and image benchmarks covering reasoning and referring segmentation tasks.}
    \label{fig:Radar}
\end{figure}

\begin{figure*}[t!]
    \centering
    \includegraphics[width=0.90\textwidth]{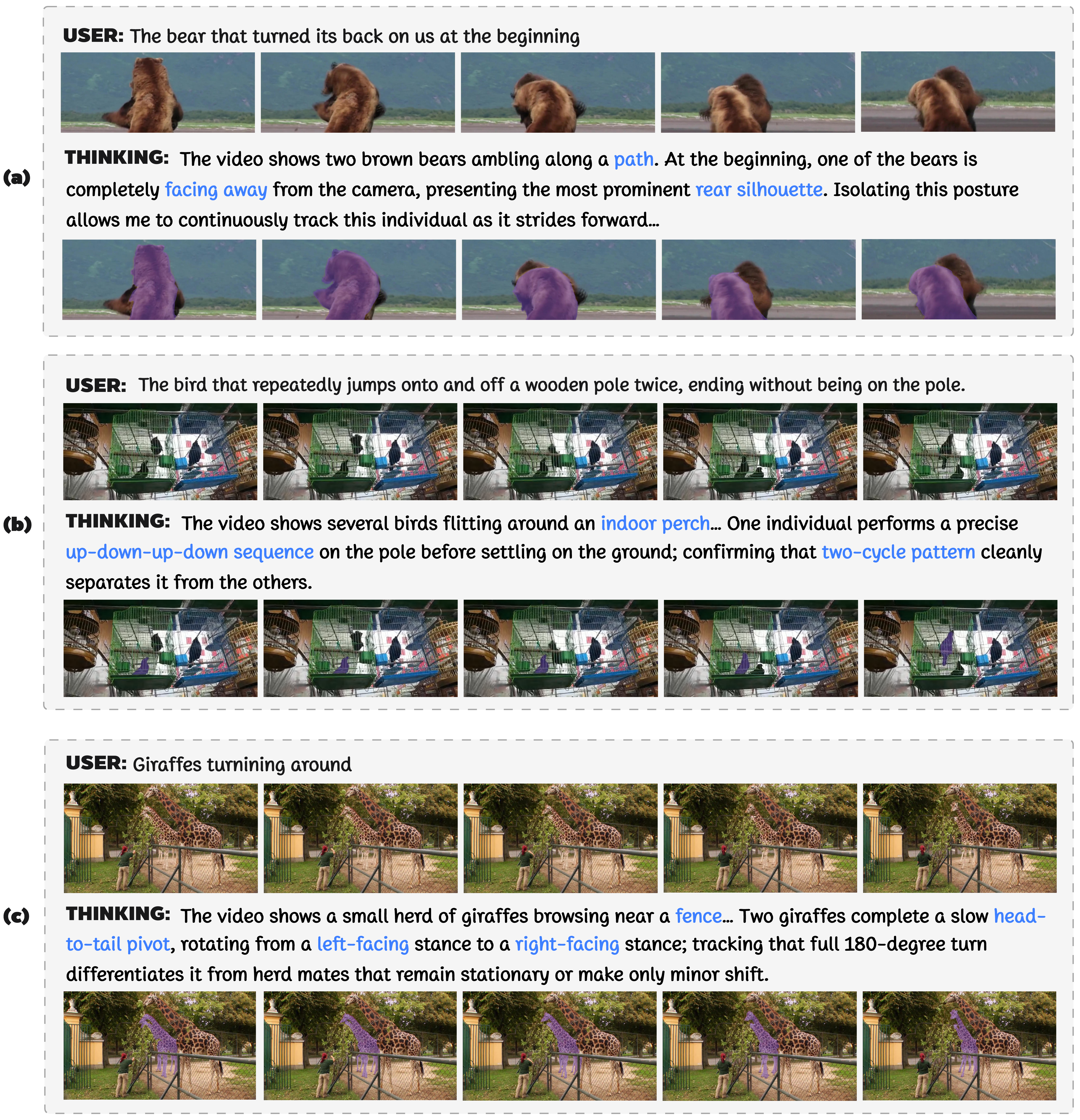}
    \caption{Our VideoSeg-R1 effectively segments and tracks in challenging scenarios, including: (a) objects in crowded scenes; (b) multiple objects with rapid motion; and (c) diverse targets appearing simultaneously.}
    \label{fig:big_pic}
\end{figure*}

We bridge this gap with \textbf{VideoSeg-R1}—the first framework that
casts RVOS as an \emph{RL-driven, reasoning-centric} problem,
capable of handling multi-target queries. VideoSeg-R1 adopts a \emph{decoupled} three-stage design:
\begin{enumerate}[label=(\roman*),nosep]
   \item \textit{Hierarchical text-guided frame sampling} progressively narrows the search space, mimicking human coarse-to-fine attention to effectively isolate key clips while reducing redundancy.

  \item \textit{GRPO-enhanced multimodal reasoning} operates on the
        selected frames, generating explicit spatial cues
        (bboxes \& points) together with a chain of thought whose length
        is modulated by a \emph{task-difficulty-aware soft penalty}.
  \item \textit{Seg-prop decoupling} sends the sparse cues to
        state-of-the-art segmentation (\textsc{SAM2}) and bidirectional
        propagation (XMem) modules, producing
        pixel-accurate masks for every frame at a fraction of the
        computation cost.
\end{enumerate}

\textbf{Extensive experiments} on Ref-YouTube-VOS~\cite{seo2020urvos}, MeViS~\cite{dingMeViSLargescaleBenchmark2023}, Ref-DAVIS17~\cite{pont-tuset2017DAVISChallenge2018} and ReVOS~\cite{yanVISAReasoningVideo2024} show that VideoSeg-R1 sets a new state of the art,
with particularly large gains (\(\!\ge6.0\%\) J\&F) on
reasoning-intensive queries.
Ablations confirm that (i)~hierarchical sampling improves performance by precisely locating key frames, (ii)~difficulty-aware GRPO boosts reasoning fidelity,
and (iii)~decoupling reasoning from propagation is essential for
temporal stability.

Our Contributions are listed below:
\begin{itemize}[leftmargin=1.5em,nosep]
  \item We introduce \textbf{the first RL framework for
        reasoning-aware RVOS}, unifying explicit chain-of-thought
        generation with temporal mask propagation.
  \item We devise a \textbf{hierarchical frame sampler} that aligns
        computational effort with query semantics, enabling efficient
        processing of minute-long videos.
  \item We propose a \textbf{task-difficulty-aware soft length penalty}
        that adaptively controls reasoning depth under GRPO,
        improving both accuracy and efficiency.
  \item We achieve new \textbf{state-of-the-art} results on multiple benchmarks, validating the effectiveness and generality of
        VideoSeg-R1.
\end{itemize}

By marrying reinforcement learning with video segmentation,
VideoSeg-R1 opens a new research avenue for \emph{explicitly
interpretable, resource-aware} video understanding.

\section{Related Work}


%

\subsection{Multi-Modal Large Language Model}



Multimodal Large Language Models (MLLMs) have significantly advanced vision-language tasks in recent years, with notable examples including InstructBLIP~\cite{dai2023instructblip}, InternGPT~\cite{liuInternGPTSolvingVisionCentric2023}, QwenVL~\cite{baiQwen25VLTechnicalReport2025}, and Intern-Video2~\cite{wangInternVideo2ScalingFoundation2024}. Despite their success, existing MLLMs still have considerable scope for improvement in reasoning abilities. To enhance these capabilities, methods such as process-based reward models~\cite{lightman2023let,uesato2022solving}, reinforcement learning ~\cite{kumar2024training} and search algorithms~\cite{feng2023alphazero,trinh2024solving} have been explored. Among these, DeepSeek R1, trained with the GRPO algorithm, has demonstrated strong reasoning performance and test-time scalability. Building on this, reinforcement learning techniques have recently been applied to multimodal large language models. Examples include Open-R1-Multimodal~\cite{EvolvingLMMsLab2025}, emphasizing mathematical reasoning, and R1-V~\cite{R1-V}, excelling at counting tasks. Additionally, studies such as Seg-Zero, SAM-R1, and Seg-R1 have targeted fine-grained pixel-level understanding. However, current research primarily addresses static image scenarios and lacks comprehensive temporal reasoning.

Extending these models to video domains introduces significant challenges in managing temporal dimensions. Issues like long-term video perception, language ambiguity, object occlusion, rapid motion, and appearance variations complicate temporal reasoning. To bridge this gap, we propose VideoSeg-R1, which integrates reinforcement learning into the Reasoning Video Object Segmentation task for the first time, significantly enhancing pixel-level temporal reasoning capabilities in video scenarios.

\subsection{MLLMs for Segmentation}
Early methods like LISA~\cite{laiLISAReasoningSegmentation2024} introduced a special \texttt{<SEG>} token, bridging MLLMs with segmentation models such as SAM to produce accurate segmentation masks. Following this paradigm, methods like PixelLM~\cite{renPixelLMPixelReasoning2024}, GLaMM~\cite{rasheed2024glamm}, VisionLLM~\cite{wang2023visionllm}, and TEXT4SEG~\cite{lan2024text4seg} focused primarily on static image segmentation, exhibiting limited adaptability for video object segmentation.

For example, TrackGPT~\cite{stroh2024trackgpt} extended LISA by updating tokens iteratively across video frames, yet ignored temporal dependencies. VISA addressed this with keyframe sampling to handle multiple frames but suffered from cumulative errors due to inaccurate keyframe selection. VideoLISA reduced computational load through sparse sampling but lacked adaptive keyframe extraction, causing redundancy. Although ViLLa improved sampling accuracy with key segment extraction, it faced significant computational overhead in long videos or lengthy action sequences. Moreover, these methods typically rely on SFT, limiting generalization to out-of-distribution (OOD) samples and causing catastrophic forgetting, hindering real-world applicability.

To address these issues, we propose VisionSeg-R1, featuring a decoupled design~\cite{guo2025decouplingcontinualsemanticsegmentation,laiLISAReasoningSegmentation2024} and a Hierarchical Text-guided Frame Sampler that mimics human attention strategies to effectively reduce redundancy. Additionally, we leverage the GRPO algorithm to enhance reasoning capabilities and generalization performance.

\section{Method}

\begin{figure*}[t!]
    \centering
    \includegraphics[width=1\linewidth]{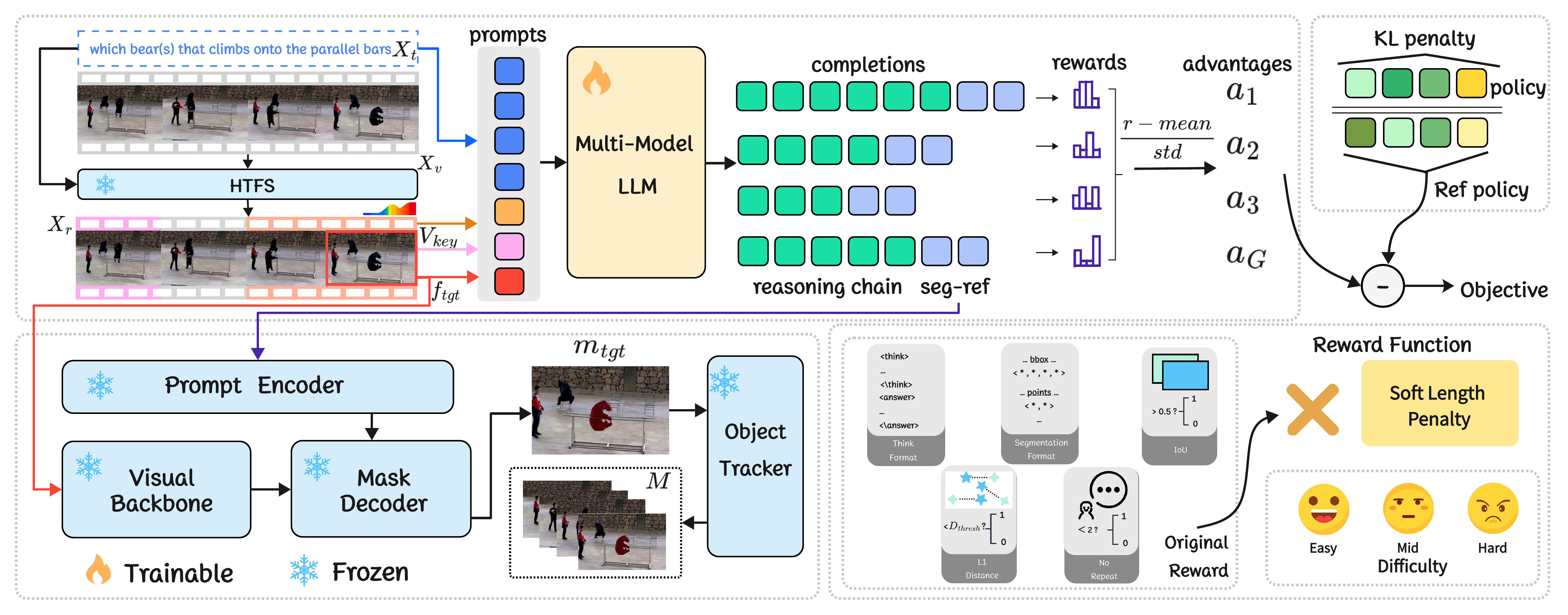} 
\caption{Overview of VideoSeg-R1, which consists of the following three stages: 
(1) a hierarchical text-guided frame sampler to emulate human attention; 
(2) a reasoning model that produces spatial cues along with explicit reasoning chains; and 
(3) a segmentation-propagation stage using SAM2 and XMem.}

    \label{fig:pdf-insert}
    \vspace{-13pt}
\end{figure*}

\subsection{Pipeline Formulation} Given a text query instruction $x_t$ and an input video $x_v = \{f_t\}_{t=1}^T \in \mathbb{R}^{T \times H \times W \times 3}$ with $T$ frames, our goal is to design a model $\phi_\theta(\cdot)$ that generates a binary segmentation mask sequence $M = \{m_t\}_{t=1}^T \in \mathbb{R}^{T \times H \times W}$, which precisely localizes the target object in the video based on the query semantics.
\begin{equation}
M = \phi_\theta(x_t, x_v) 
\end{equation}
The complexity of the text query $x_t$ varies: it may be a simple phrase that directly describes the appearance, action, or position of the target (e.g., ``the woman in red”), or it could involve expressions that require world knowledge or commonsense reasoning (e.g., ``the person who looks like a doctor”). It may also require complex inference about the video content and future developments (e.g., ``who is most likely to be the main character in this wedding?”). The latter two types of queries demand more advanced semantic understanding and reasoning capabilities from the model.

To address these challenges and leverage recent advances in the reasoning abilities of LLM, we propose a reasoning-based video object segmentation pipeline. We formulate the overall task as a joint problem of referring image segmentation and video mask propagation, consisting of the following three stages: (1) A hierarchical text-guided frame sampler to emulate human attention; (2) A reasoning model that produces spatial cues along with explicit reasoning chains; and (3) A segmentation-propagation stage using SAM2 and XMem. 
\subsection{VideoSeg-R1 Model}









\noindent\textbf{Hierarchical Text-guided Frame Sampler.} To better align with human perception in long video understanding, we propose a hierarchical text-guided sampling strategy that simulates progressive attention convergence. In long video understanding, humans typically begin with a coarse estimate of when an event may occur, and progressively refine their attention to identify the precise frame. Inspired by this, we design a multi-round reasoning structure that iteratively compresses semantics to locate key segment and traget frame. 

Specifically, given an input video $x_v \in \mathbb{R}^{T \times H \times W \times 3}$ and a textual query $x_t$, we first prompt a video understanding LLM (e.g., Qwen2.5-VL~\cite{baiQwen25VLTechnicalReport2025}) to predict the time intervals semantically relevant to the query. We extract $K$ temporal intervals $(t_i^{\text{start}}, t_i^{\text{end}})$ from multiple responses and compute their average to determine the key segment boundaries $[t_{\text{key}}^{\text{start}},\ t_{\text{key}}^{\text{end}}]$. The corresponding set of frames is denoted as  $V_{\text{key}} = \{f_{t_{\text{key}}^{\text{start}}}, f_{t_{\text{key}}^{\text{start}} + 1}, \dots, f_{t_{\text{key}}^{\text{end}}}\}$
, has a length of $T_{\text{key}}$. 
To prevent errors caused by performing frame-level localization on overly long segments, we introduce a relative segment length threshold $\delta \in (0,1)$. When the key segment length $T_{\text{key}} > \delta\cdot T$, we continue applying semantic compression to $V_{\text{key}}$ until $T_{\text{key}} \leq \delta \cdot T$, indicating that the model’s attention has sufficiently converged for frame-level reasoning. In the frame-level localization phase, the model predict the approximate percentage position of the target frame within the key segment.




We collect the top-$M$ predicted values $\{p_i\}_{i=1}^{M}$ and compute the average $\bar{p}$ to determine the final target frame \(f_{\text{tgt}} = f_{t_{\text{key}}^{\text{start}} + \lfloor T_{\text{key}} \cdot \bar{p} \rfloor}\). To implement the semantic convergence process described above, we introduce two stage-specific prompt templates. See the ~\cref{sec:Prompts for Hierarchical Text-guided Frame Sampler} for details. To provide global context, we also employ an adaptive global sampling strategy to sample reference frames \(X_r\). Different sampling strategies are detailed in the \cref{sec:Reference Sampling Strategies}.


\noindent\textbf{Reasoning Model.} We adopt Qwen2.5-VL~\cite{baiQwen25VLTechnicalReport2025} as the reasoning model $F_{\text{reason}}$. At this stage, the model takes as input the key segment frames $V_{\text{key}}$, the target frame $f_{\text{tgt}}$, and the reference frame set $X_r$, along with the textual query $x_t$, and performs multimodal reasoning.

During reinforcement learning, the model is optimized to generate structured outputs. These outputs are parsed by a post-processing function $G$ to extract the target bounding box $B$, a central point $P_{\text{central}}$, and a negative point $P_{\text{neg}}$, which help improve spatial precision and distinguish between multiple objects. Each point is represented as $P = (x, y, l)$, where $(x, y)$ denotes the spatial coordinates, and $l \in \{0, 1\}$ is a binary label indicating whether the point is positive ($l = 1$) or negative ($l = 0$). This process can be formally expressed as:
\begin{equation}
B,\ P_{\text{central}},\ P_{\text{neg}} = G(F_{\text{reason}}(V_{\text{key}},\ f_{\text{tgt}},\ X_r,\ x_t))
\end{equation}

\noindent\textbf{Segmentation and propagation stage.}
We employ SAM2~\cite{ravi2024sam} as the segmentation model $F_{\text{seg}}$, chosen for its high accuracy and efficient inference. Given the target frame $f_{\text{tgt}}$, a visual backbone $E_v$ first extracts its features. The predicted bounding box $B$, together with a positive point $P_{\text{central}}$ and a negative point $P_{\text{neg}}$, serves as spatial prompts to guide the segmentation model in generating the target mask:
\begin{equation}
 m_{\text{tgt}} = \text{SAM2}(E_v(f_{\text{tgt}}), B, P_{\text{central}}, P_{\text{neg}})   
\end{equation}

To extend the segmentation across the entire video, we apply XMem~\cite{cheng2022xmem}, an advanced object tracking model, to propagate $m_{\text{tgt}}$ bidirectionally: 
\begin{equation}
    M = \{ m_t \}_{t=1}^{T} = \text{OT}(m_{\text{tgt}}, x_v) 
\end{equation}

\subsection{Reward Functions}

\noindent\textbf{Original Reward.} We adopt the original reward design in VisionReasoner~\cite{liu2025visionreasonerunifiedvisualperception}. The overall reward function consists of the following components:
\begin{equation}
R_{\text{original}} = R_{\text{format}} + R_{\text{seg\_accuracy}} + R_{\text{non\_repeat}}
\end{equation}
\begin{equation}
R_{\text{format}} = R_{\text{reason\_format}} + R_{\text{seg\_format}}
\end{equation}
which assess the reasoning format, segmentation format, segmentation accuracy, and non-redundant reasoning, respectively. $R_{\text{non\_repeat}}$ measures the diversity of the reasoning process by assigning higher rewards to outputs composed of unique, non-repetitive sentences. The segmentation accuracy reward $R_{\text{seg\_accuracy}}$ comprises the evaluation of mask IoU, point-level L1 distance, and bounding box-level L1 distance. See Appendix~\ref{sec:Reward Functions} for details.

\noindent\textbf{Negative Point Distance Reward.} 
To enhance the model’s ability to distinguish foreground from background, we introduce a negative point distance reward.
For each predicted negative point $(x, y, 0)$, we compute its minimum L1 distance to all ground-truth target regions (annotated masks).
If the distance is positive and does not exceed the predefined threshold $\tau_{\text{neg}}$ (40 pixels), the point is considered valid and receives a reward increment of $\frac{1}{K}$, where $K$ is the number of predicted negative points;
if the distance is zero or negative—that is, if the point lies inside or on the boundary of the target mask—no reward is given.
This mechanism encourages the model to place negative points close to, but outside, the target regions, thereby improving foreground–background separability.

\noindent\textbf{Task Difficulty.}
To enable efficient training for video reasoning segmentation, we estimate an instance-level task difficulty score $D \in [1, 10]$ for each sample. Specifically, we prompt a MLLM to rate the sample across five dimensions: \textit{scene complexity}, \textit{segmentation challenge}, \textit{temporal ambiguity}, \textit{motion complexity}, and \textit{linguistic ambiguity}. Each dimension is scored on a 1--10 scale, and the final difficulty score is computed as the average of these five ratings. We also categorize difficulty into three levels (easy, medium, and hard) using thresholds $\tau_{\text{easy}}$ and $\tau_{\text{hard}}$. This difficulty prior is then used to guide reasoning token budget allocation in reinforcement learning, allowing the model to adaptively adjust reasoning length based on the sample's difficulty. Details are provided in the \cref{sec:Task Difficulty Scoring Details}.

\noindent\textbf{Soft Length Penalty.} To enable adaptive control over reasoning length under varying task complexities, we propose a soft length penalty mechanism. Unlike traditional methods that rely on hard truncation, our approach encourages concise reasoning for simple tasks while allowing more detailed reasoning for complex ones. Specifically, we define the expected reasoning token budget \(L_{\text{budget}}\) based on the task difficulty score \(D\) as follows:
\(L_{\text{budget}} =L(D)\), where \(L(D)\) denotes the base budget allocated according to task difficulty \(D\); the detailed mapping strategy is provided in \cref{sec:budget mapping}. Let \(L_{\text{used}}\) be the number of reasoning tokens actually generated by the model. When it exceeds the budget, a soft penalty is applied to the reward:
\begin{equation}
s(L_{\text{used}}, L_{\text{budget}}) = \begin{cases}
1 - \beta \cdot (L_{\text{used}} - L_{\text{budget}}), \\
\quad\quad\quad\quad \text{if } L_{\text{used}} > L_{\text{budget}} \\
1, \quad \text{otherwise}
\end{cases}
\end{equation}
The final reward is computed as:
\begin{equation}
 R_{\text{final}} = R_{\text{original}} \cdot s(L_{\text{used}}, L_{\text{budget}})   
\end{equation}
\subsection{Multi-object Matching}
To support multi-object segmentation, our framework employs batched computation and the Hungarian algorithm to effectively handle the many-to-many matching problem under bounding box IoU reward, bounding box L1 reward, and center point L1 reward. For each observed object $o_j$, we maintain a list of predicted bounding boxes $(B_{\text{pred}}^i)_{i=1}^K$ and predicted center points $(P_{\text{pred}}^i)_{i=1}^K$, and compute the reward scores in batch with respect to the ground-truth bounding boxes $(B_{\text{GT}}^i)_{i=1}^N$ and ground-truth center points $(P_{\text{GT}}^i)_{i=1}^N$. Subsequently, we use the Hungarian algorithm to compute the optimal one-to-one assignment. This design ensures both optimal alignment between predictions and ground-truth annotations and efficient computation performance.



\section{Experiment}
\begin{table*}[t]
\centering

\begin{tabular}{l|ccc|ccc|ccc}
\toprule
\multirow{2}{*}{\textbf{Method}} & \multicolumn{3}{c|}{\textbf{Ref-YouTube-VOS}} & \multicolumn{3}{c|}{\textbf{Ref-DAVIS17}} & \multicolumn{3}{c}{\textbf{MeViS}} \\
 & $\mathcal{J}\&\mathcal{F}$ & $\mathcal{J}$ & $\mathcal{F}$ & $\mathcal{J}\&\mathcal{F}$ & $\mathcal{J}$ & $\mathcal{F}$ & $\mathcal{J}\&\mathcal{F}$ & $\mathcal{J}$ & $\mathcal{F}$ \\
\midrule
URVOS~\cite{seo2020urvos}             & 47.2 & 45.3 & 49.2 & 51.6 & 47.3 & 56.0 & 27.8 & 25.7 & 29.9 \\
MTTR~\cite{mttr}                              & 55.3 & 54.0 & 56.6 & -    & -    & -    & 30.0 & 28.8 & 31.2 \\
LBDT~\cite{ding2022language}               & 49.4 & 48.2 & 50.6 & 54.1 & -    & -    & 29.3 & 27.8 & 30.8 \\
ReferFormer~\cite{wuLanguageQueriesReferring2022} & 62.9 & 61.3 & 64.1 & 61.1 & 58.1 & 64.1 & 31.0 & 29.8 & 32.2 \\
LMPM~\cite{dingMeViSLargescaleBenchmark2023}               & -    & -    & -    & -    & -    & -    & 37.2 & 34.2 & 40.2 \\
OnlineRefer~\cite{wuOnlineReferSimpleOnline2023} & 62.9 & 61.0 & 64.7 & 62.4 & 59.1 & 65.6 & -    & -    & -    \\
DsHmp~\cite{heDecouplingStaticHierarchical2024}             & 67.1 & 65.0 & 69.1 & 64.9 & 61.7 & 68.1 & 46.4 & 43.0 & 49.8 \\
TrackGPT~\cite{TrackGPT}                      & 59.5 & 58.1 & 60.8 & 66.5 & 62.7 & 70.4 & 41.2 & 39.2 & 43.1 \\
\midrule
LISA~\cite{laiLISAReasoningSegmentation2024}               & 54.4 & 54.0 & 54.8 & 66.0 & 63.2 & 68.8 & 37.9 & 35.8 & 40.0 \\
PixelLM~\cite{renPixelLMPixelReasoning2024}         & 55.0 & 54.3 & 55.7 & 66.7 & 63.4 & 70.0 & 38.7 & 36.3 & 41.1 \\
VideoLISA~\cite{fu2025video}     & 61.7 & 60.2 & 63.1 & 67.7 & 63.8 & 71.5 & 42.3 & 39.4 & 45.2 \\
VISA~\cite{yanVISAReasoningVideo2024}               & 63.0 & 61.4 & 64.6 & 70.4 & 66.7 & 73.8 & 44.5 & 41.8 & 47.1 \\
ViLLa ~\cite{zhengViLLaVideoReasoning2025}& 73.3 & 70.5 & 76.6 & 74.3 & 70.6 & 78.0 & 49.4 & 46.5 & 52.3 \\

\rowcolor{gray!20}
\textbf{VideoSeg-R1(Qwen2.5-VL-3B)} & \textbf{75.9} & \textbf{73.3} & \textbf{78.5} & \textbf{77.5} & \textbf{74.4} & \textbf{80.5} & \textbf{53.0} & \textbf{50.7} & \textbf{55.3} \\

\rowcolor{gray!20}
\textbf{VideoSeg-R1(Qwen2.5-VL-7B)}                 & \textbf{81.3} & \textbf{78.2} & \textbf{84.4} & \textbf{79.8} & \textbf{77.4} & \textbf{82.2} & \textbf{55.3} & \textbf{52.7} & \textbf{57.8} \\
\bottomrule
\end{tabular}

\caption{Referring Video Object Segmentation on Ref-YouTube-VOS, Ref-DAVIS17, and MeViS.}
\label{tab:refvos_results}
\end{table*}

\begin{table*}[ht]
\centering

\label{tab:segmentation_comparison}
\resizebox{\textwidth}{!}{
\begin{tabular}{l| l|ccc|ccc|cc|cc}
\toprule
\multirow{2}{*}{\textbf{Methods}} & \multirow{2}{*}{\textbf{Backbone}} & \multicolumn{3}{c|}{\textbf{refCOCO}} & \multicolumn{3}{c|}{\textbf{refCOCO+}} & \multicolumn{2}{c|}{\textbf{refCOCOg}} & \multicolumn{2}{c}{\textbf{ReasonSeg}} \\
& & val & testA & testB & val & testA & testB & val(U) & test(U) & gIoU & cIoU \\
\midrule
MCN~\cite{luoMultiTaskCollaborativeNetwork2020} & Darknet53 & 62.4 & 64.2 & 59.7 & 50.6 & 55.0 & 44.7 & 49.2 & 49.4 & - & - \\
VLT~\cite{dingVisionLanguageTransformerQuery2021} & Darknet53 & 65.7 & 68.3 & 62.7 & 55.5 & 59.2 & 49.4 & 53.0 & 56.7 & - & - \\
CRIS~\cite{wangCRISCLIPDrivenReferring2022} & ResNet101 & 70.5 & 73.2 & 66.1 & 62.3 & 68.1 & 53.7 & 59.9 & 60.4 & - & - \\
LAVT~\cite{yangLAVTLanguageAwareVision2022} & Swin-B & 72.7 & 75.8 & 68.8 & 62.1 & 68.4 & 55.1 & 61.2 & 62.1 & - & - \\
ReLA~\cite{liuGRESGeneralizedReferring2023} & Swin-B & 73.8 & 76.5 & 70.2 & 66.0 & 71.0 & 57.7 & 65.0 & 66.0 & - & - \\
X-Decoder~\cite{zou2023generalized} & DaViT-L & - & - & - & - & - & - & 64.6 & - & 22.6 & 17.9 \\
SEEM~\cite{zou2023segment} & DaViT-L & - & - & - & - & - & - & 65.7 & - & 25.5 & 21.2 \\
\midrule
LISA~\cite{laiLISAReasoningSegmentation2024} & LLaVA-7B & 74.9 & 79.1 & 72.3 & 65.1 & 70.8 & 58.1 & 67.9 & 70.6 & 52.9 & 54.0 \\
VISA~\cite{yanVISAReasoningVideo2024}  & Chat-UniVi-7B & 72.4 & 75.5 & 68.1 & 59.8 & 64.8 & 53.1 & 65.5 & 66.4 & 52.7 & 57.8 \\
VideoLISA~\cite{fu2025video} & LLaVA-Phi-3-V-3.8B & 73.8 & 76.6 & 68.8 & 63.4 & 68.8 & 56.2 & 68.3 & 68.8 & 61.4 & 67.1 \\
\rowcolor{gray!20}
\textbf{VideoSeg-R1 (Ours)} & \textbf{Qwen2.5-VL-3B}& \textbf{75.1} & \textbf{79.2} & \textbf{72.8} & \textbf{67.2} & \textbf{72.8} & \textbf{59.9} & \textbf{69.7} & \textbf{71.0} & \textbf{65.1} & \textbf{63.7}
\\
\rowcolor{gray!20}
\textbf{VideoSeg-R1 (Ours)} & \textbf{Qwen2.5-VL-7B}& \textbf{78.2} & \textbf{82.3} & \textbf{75.1} & \textbf{71.8} & \textbf{76.1} & \textbf{64.7} & \textbf{73.1} & \textbf{74.1} & \textbf{69.8} & \textbf{68.2}
\\
\bottomrule
\end{tabular}
}

\caption{Referring image segmentation on the refCOCO, refCOCO+, refCOCOg, and ReasonSeg datasets.}
\label{tab:image_eval}
\end{table*}

\begin{table}[ht]
\centering

\begin{tabular}{lccc}
\toprule
\textbf{Method} & \multicolumn{3}{c}{\textbf{ReVOS}} \\
\cmidrule(lr){2-4}
 & $\mathcal{J\&F}$ & $\mathcal{J}$ & $\mathcal{F}$ \\
\midrule
LISA-LLAVA-13B & 41.8 & 39.6 & 43.9 \\
VISA-Chat-UniVi-13B & 50.9 & 48.8 & 52.9 \\
VISA-InternVideo2-6B & 52.4 & 50.1 & 54.7 \\

ViLLa-InternVideo2-6B& 57.0 & 54.9 & 59.1 \\
\rowcolor{gray!20}
VideoSeg-R1-Qwen2.5-VL-3B  & \textbf{58.6} & \textbf{56.4} & \textbf{60.8} \\
\rowcolor{gray!20}
VideoSeg-R1-Qwen2.5-VL-7B  & \textbf{61.1} & \textbf{58.2} & \textbf{64.0} \\
\bottomrule
\end{tabular}
\caption{Video Reasoning Segmentation on ReVOS.}
\label{tab:ReVOS}
\end{table}

\begin{table}[ht]
\centering
\resizebox{\linewidth}{!}{%
\begin{tabular}{l|c|c|c|c|c}
\toprule
\multirow{2}{*}{\textbf{Model}} & \multirow{2}{*}{\textbf{Type}} & \multirow{2}{*}{\textbf{CoT}} &
\multicolumn{1}{c|}{\textbf{MeViS}} & \multicolumn{1}{c|}{\textbf{ReVOS}} & \multicolumn{1}{c}{\textbf{ReasonSeg}} \\
 &  &  & J\&F & J\&F & gIoU \\
\midrule
Baseline         &    &      & 38.1 & 40.2 & 56.1 \\
VideoSeg-R1      & SFT  & \xmark & 45.4 & 50.9 & 59.4 \\
VideoSeg-R1      & RL   & \xmark & 52.6 & 58.9 & 67.2 \\
VideoSeg-R1      & RL   & \cmark & \textbf{55.3} & \textbf{61.1} & \textbf{69.8} \\
\bottomrule
\end{tabular}
}

\caption{Performance comparison between SFT and RL.}
\label{tab:sft_vs_rl}
\vspace{-13pt}
\end{table}


\begin{table}[ht]
\centering
\begin{tabular}{c c c c c}
\toprule
\textbf{ID} & \textbf{Key Segment} & \textbf{Target Frame} & \textbf{MeViS} & \textbf{ReVOS} \\
\midrule
1 & \xmark & \xmark & 45.7 & 53.9 \\
2 & \xmark & \cmark & 52.5 & 57.6 \\
3 & \cmark & \xmark & 49.2 & 56.3 \\
4 & \cmark & \cmark & \textbf{55.3} & \textbf{61.1} \\
\bottomrule
\end{tabular}
\caption{Ablation study of target localization strategies.}
\label{tab:tfs_ablation}
\end{table}

\begin{table}[ht]
\centering
\resizebox{\linewidth}{!}{%
\begin{tabular}{l|c|c|c|c|c}
\toprule
\multirow{2}{*}{\textbf{Model}} & \textbf{Bbox} & \textbf{$P_{\text{central}}$ }& \textbf{$P_{\text{neg}}$} & \textbf{ReVOS} & \textbf{ReasonSeg} \\
 &  &  &  & J\&F & gIoU \\
\midrule
Baseline         &             &                &              & 40.2 & 56.1 \\
VideoSeg-R1      & \xmark      & \cmark         & \xmark       & 56.5 & 64.0 \\
VideoSeg-R1      & \cmark      & \xmark         & \xmark       & 57.2 & 65.7 \\
VideoSeg-R1      & \cmark      & \cmark         & \xmark       & 60.3 & 68.2 \\
VideoSeg-R1      & \cmark      & \xmark         & \cmark       & 58.1 & 66.5 \\
VideoSeg-R1      & \cmark      & \cmark         & \cmark       & \textbf{61.1} & \textbf{69.8} \\
\bottomrule
\end{tabular}
}
\caption{Ablation study on the effect of Bbox, $P_{\text{central}}$, and $P_{\text{neg}}$ as spatial prompts. }
\label{tab:ablation_bbox_points}
\vspace{-12pt}
\end{table}

\begin{table}[ht]
\centering
\begin{tabular}{l|cc|cc}
\toprule
\multirow{2}{*}{\textbf{Dataset}} 
& \multicolumn{2}{c|}{\textbf{$\mathcal{J}\&\mathcal{F}\uparrow$}} 
& \multicolumn{2}{c}{\textbf{\#Token$\downarrow$}} \\
& w/ & w/o & w/ & w/o \\
\midrule
Ref-YouTube-VOS & 81.3 & 78.4 & 43.2 & 77.1 \\
Ref-DAVIS17     & 79.8 & 75.2 & 42.8 & 62.4 \\
MeViS           & 55.3 & 52.2 & 52.7 & 82.3 \\
ReVOS           & 61.1 & 57.3 & 56.1 & 89.9 \\
\bottomrule
\end{tabular}
\caption{Ablation study of Soft Length Penalty.}
\label{tab:soft_length_penalty}
\vspace{-10pt}
\end{table}

\subsection{Dataset}
\noindent\textbf{Training Dataset.} We train VisionSeg-R1 using the Ref-YouTube-VOS~\cite{seo2020urvos}, MeViS~\cite{dingMeViSLargescaleBenchmark2023} and Ref-DAVIS17~\cite{pont-tuset2017DAVISChallenge2018}. For the mask annotations in Referring VOS datasets, we extract the leftmost, topmost, rightmost, and bottommost pixels of each target mask to generate the bounding box \( B \). In addition, we compute the center point coordinates of each mask. To support multi-object expressions, we handle multiple objects per image by: (i) using one center point per object (ii) assembling all corresponding bounding boxes and center points into respective lists.

\noindent\textbf{Evaluation Dataset.} For evaluation, we test the model on both video and image datasets: (1) For video datasets, we use the ReVOS~\cite{yanVISAReasoningVideo2024} dataset to evaluate the performance of ReasonVOS, and the Ref-YouTube-VOS, MeViS and Ref-DAVIS17 to evaluate the performance of vanilla Referring VOS performance. 
(2) For image datasets, we use ReasonSeg~\cite{laiLISAReasoningSegmentation2024}, refCOCO~\cite{kazemzadehReferItGameReferringObjects2014}, refCOCO+~\cite{kazemzadehReferItGameReferringObjects2014}, and refCOCOg~\cite{maoGenerationComprehensionUnambiguous2016} 
to evaluate the generalization ability of the VisionSeg-R1 model on image-level segmentation tasks.
\subsection{Implementation Details}
We adopt Qwen2.5-VL-7B and Qwen2.5-VL-3B~\cite{baiQwen25VLTechnicalReport2025} as our reasoning models and video understanding models, and use SAM2-Large~\cite{ravi2024sam} as the default segmentation model. In addition, we utilize XMem~\cite{cheng2022xmem}, a semi-supervised video object segmentation method, as the object tracker. We train VisionSeg-R1 using the DeepSpeed~\cite{rasley2020deepspeed} library on  8 NVIDIA 80G A100 GPUs. During training, the Hierarchical Text-guided Frame Sampler, the visual backbone, the SAM2 decoder, the prompt encoder, and the object tracker are all frozen. Only the multi-modal LLM is updated. Note that the Hierarchical Text-guided Frame Sampler is only used during inference. During training, we use a total batch size of 16 with a sampling number of 8 per training step. The initial learning rate is set to 1e-6 and the weight decay is 0.01. More training details are provided in the \cref{sec:Implementation Details}.

\subsection{Evaluation Metrics}
For image-based evaluation, we adopt two commonly used metrics: gIoU and cIoU, following prior works ~\cite{kazemzadehReferItGameReferringObjects2014,laiLISAReasoningSegmentation2024}. Specifically, gIoU is computed as the average of all per-image Intersection-over-Unions (IoUs), while cIoU is defined by the cumulative intersection over the cumulative union. For video-based evaluation, we follow previous studies~\cite{wuLanguageQueriesReferring2022, wuOnlineReferSimpleOnline2023}, using region similarity (J), contour accuracy (F), and their average value (J\&F).

\subsection{Comparison}
We compare our model with prior works through quantitative evaluations on standard benchmarks (Tables~\ref{tab:refvos_results}, \ref{tab:image_eval}, \ref{tab:ReVOS}) and qualitative comparisons provided in the \cref{sec:Qualitative Results}.

\noindent\textbf{Referring VOS.} In the task of referring video object segmentation, VideoSeg-R1 achieves leading performance on three standard benchmarks: Ref-YouTube-VOS, Ref-DAVIS17, and MeViS, with J\&F scores of 81.3, 79.8, and 55.3 respectively. These results significantly surpass mainstream methods such as ViLLa (73.3, 74.3, 49.4) and VISA (63.0, 70.4, 44.5), as shown in Table~\ref{tab:refvos_results}.

\noindent \textbf{Image Datasets.}
Images can be treated as single-frame videos, allowing VideoSeg-R1 to be directly applied to image datasets without any modification. As shown in Table~\ref{tab:image_eval}, our method consistently outperforms existing state-of-the-art approaches across all benchmarks. On refCOCO, VideoSeg-R1 achieves a validation score of 78.2, surpassing LISA and VideoLISA by 3.3 and 4.4 points, respectively. For refCOCO+ and refCOCOg, our model obtains scores of 71.8 and 73.1, outperforming LISA by 6.7 and 5.2 points, respectively. Most notably, on the reasoning-intensive ReasonSeg dataset, VideoSeg-R1 achieves 69.8 gIoU and 68.2 cIoU, outperforming VideoLISA by 8.4 and 1.1 points, respectively. These results clearly demonstrate the superior performance of our model in handling both standard and reasoning-based referring image segmentation tasks.

\noindent \textbf{ReVOS.} The results comparison on ReVOS are shown in Table~\ref{tab:ReVOS}. VideoSeg-R1-Qwen2.5-VL-7B achieves a $\mathcal{J\&F}$ score of 61.1, outperforming the best existing SFT method, ViLLa-InternVideo2-6B, by 4.1 points. Specifically, it improves $\mathcal{J}$ by 3.3 points and $\mathcal{F}$ by 4.9 points. Notably, even with only 3B parameters, VideoSeg-R1-Qwen2.5-VL-3B achieves a $\mathcal{J\&F}$ of 58.6, surpassing larger models such as ViLLa (57.0) and VISA-Chat-UniVi-13B (50.9). These results verify the significant advantage of our reinforcement learning training strategy in understanding complex language expressions and reasoning video object segmentation.

\subsection{Ablation Studies}
\noindent\textbf{SFT vs. RL.}  
We compare the performance of SFT and RL. The baseline model is Qwen2.5-VL-7B combined with SAM2-Large. In the non-CoT setting, the thinking format reward is removed, and the model no longer generates a reasoning chain. As shown in Table~\ref{tab:sft_vs_rl}, the SFT model performs reasonably well on in-domain data, but drops significantly on OOD datasets such as ReVOS and ReasonSeg, revealing limitations in world knowledge and multi-step reasoning. In contrast, the RL model achieves better results on both in-domain and OOD tasks, demonstrating stronger generalization. Moreover, introducing CoT rewards further improves performance. Compared to RL without CoT, RL+CoT achieves a 2.2-point gain in J\&F on ReVOS and a 2.6-point improvement in gIoU on ReasonSeg, showing that the reasoning process effectively enhances the model’s ability to handle OOD samples.

\noindent\textbf{Hierarchical Text-guided Frame Sampler.}  
We compare four frame selection strategies: (1) directly using the first frame (\(f_0\)); (2) directly locating the target frame; (3) randomly selecting a frame from the key segment; and (4) our proposed Hierarchical Text-guided Frame Sampler (HTFS). As shown in Table~\ref{tab:tfs_ablation}, HTFS achieves the best performance in both J\&F and gIoU metrics, significantly outperforming the other strategies. This method simulates the human attention process that progressively shifts from coarse perception to precise focus, effectively enhancing the model’s ability to locate key frames and thereby improving overall performance.

\noindent\textbf{Spatial Prompts.}
Table~\ref{tab:ablation_bbox_points} shows that using only Bbox significantly improves performance (J\&F from 40.2 to 57.2 on ReVOS, gIoU from 56.1 to 65.7 on ReasonSeg). Adding $P_{\text{central}}$ or $P_{\text{neg}}$ further boosts results, highlighting their roles in localization and foreground-background discrimination. Combining all three yields the best performance (61.1 J\&F, 69.8 gIoU), confirming their complementarity.

\noindent\textbf{Soft Length Penalty.}
As shown in Table~\ref{tab:soft_length_penalty}, introducing the Soft Length Penalty consistently improves segmentation performance across all benchmarks while significantly reducing the number of reasoning tokens. On Ref-YouTube-VOS and Ref-DAVIS17, the $\mathcal{J}\&\mathcal{F}$ scores increase by 2.9 and 4.6 respectively, with average token reductions of approximately 34 and 20. For the more challenging MeViS and ReVOS datasets, our method achieves improvements of 3.1 and 3.8 in $\mathcal{J}\&\mathcal{F}$, while reducing token usage by around 30 and 34. These results validate the effectiveness of adaptively controlling reasoning length based on task complexity.
\section{Conclusion}
We propose \textbf{VideoSeg-R1}, the first framework that introduces reinforcement learning into video reasoning segmentation. By combining hierarchical frame sampling, explicit reasoning, and decoupled segmentation-propagation, our method achieves state-of-the-art performance on multiple benchmarks. Despite strong performance, the multi-stage design and large models incur high computational cost, limiting real-time deployment. Future work will explore model simplification and tighter integration to improve practicality and scalability.

\bibliography{aaai2026}

\clearpage
\setcounter{page}{1}
\setcounter{table}{0}
\setcounter{figure}{0}
\setcounter{algorithm}{0}

\renewcommand{\thefigure}{\thesection\arabic{figure}}
\renewcommand{\thetable}{\thesection\arabic{table}}
\renewcommand{\thealgorithm}{\thesection\arabic{algorithm}}

\appendix
\setcounter{secnumdepth}{2}
\renewcommand{\thesection}{\Alph{section}}

\section{Prompts for Hierarchical Text-guided Frame Sampler}
\label{sec:Prompts for Hierarchical Text-guided Frame Sampler}
We design two stage-specific prompt templates to guide frame sampling: a coarse prompt locates key segments using time ranges, while a fine-grained prompt refines the search to a specific frame via percentage estimation. This mimics human attention for efficient temporal localization.

\begin{figure}[h]
    \centering
    \includegraphics[width=0.45\textwidth]{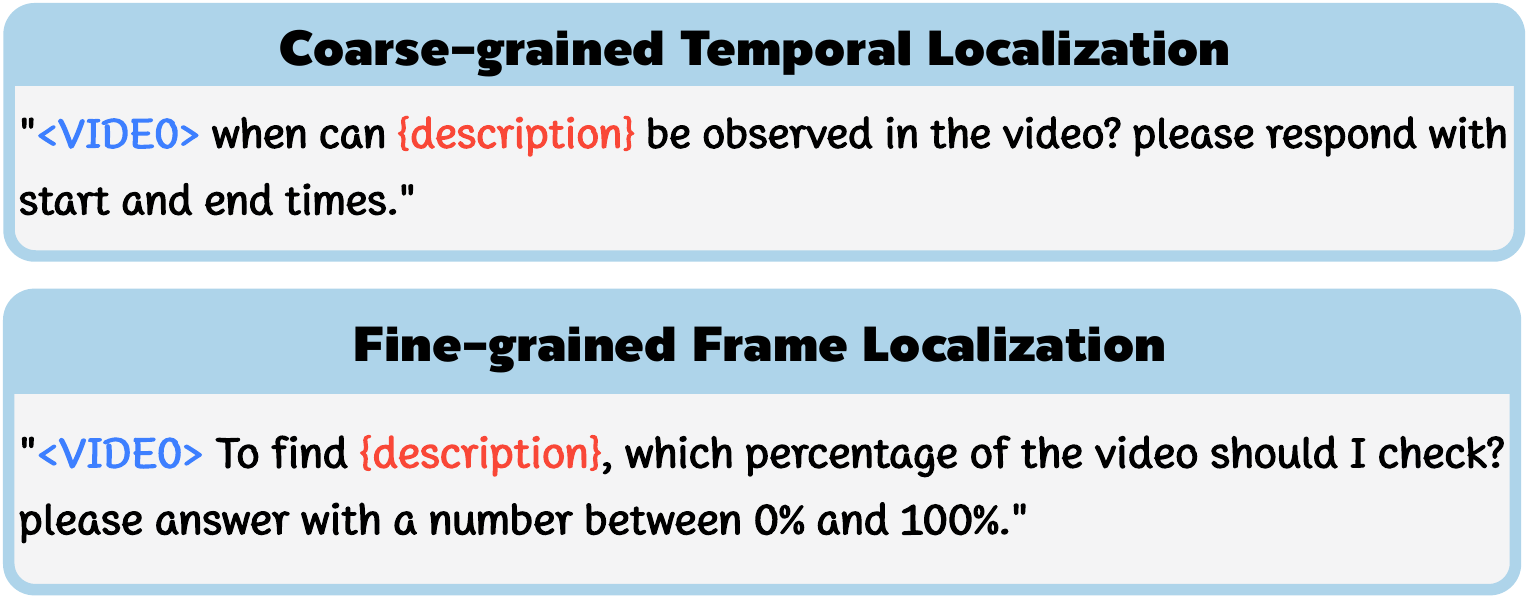}
    \caption{Prompt templates for Hierarchical Text-guided Frame Sampling. The coarse-grained temporal localization prompt guides the model to identify key segment in the video, while the fine-grained frame localization prompt instructs the model to pinpoint traget frame via percentage estimation.}
    \label{fig:enter-label}
\end{figure}

\section{Sampling Strategy}
\label{sec:Reference Sampling Strategies}
We adopt an adaptive global sampling strategy to select $T_{\text{ref}}$ reference frames for multimodal reasoning. Specifically, we design three sampling methods: global sampling, which uniformly selects $T_{\text{ref}}$ frames from the entire video excluding the key segment; local sampling, which selects reference frames from the 1 to 3 frames immediately before and after the key segment; and adaptive sampling, which combines the strengths of both approaches by selecting $1/3 $ frames from the whole video and $ 2/3 $ frames from the local regions. As shown in Table \ref{tab:sampling_ablation}, the adaptive strategy slightly outperforms individual sampling methods. Moreover, as the number of reference frames increases, the model performance consistently improves, demonstrating the importance of properly sampling contextual frames for video reasoning segmentation.
\begin{table}[h]
\centering

\begin{tabular}{l c | c c}
\toprule
\multirow{2}{*}{\textbf{Strategy}} & \multirow{2}{*}{$T_{ref}$} 
  & \textbf{ReVOS} & \textbf{ReasonSeg} \\
 &  & J\&F & gIoU \\
\midrule
\multirow{3}{*}{Global} 
  & 0  & 59.5 & 68.0 \\
  & 6  & 60.4 & 69.2 \\
  & 12 & 60.8 & 69.5 \\
\midrule
\multirow{3}{*}{Local} 
  & 0  & 59.6 & 68.2 \\
  & 6  & 60.1 & 69.1 \\
  & 12 & 60.6 & 69.4 \\
\midrule
\multirow{3}{*}{\textbf{Adaptive}} 
  & 0  & 59.3 & 67.2 \\
  & 6  & 60.9 & 69.4 \\
  & 12 & \textbf{61.1} & \textbf{69.8} \\
\bottomrule
\end{tabular}
\caption{Ablation study on different sampling strategies.}
\label{tab:sampling_ablation}
\end{table}

\section{Original Reward }
\label{sec:Reward Functions}

Original reward functions consist of three types: \textbf{format rewards}, \textbf{accuracy rewards}, and the \textbf{non repeat reward}. These rewards jointly guide the optimization process by enhancing structural correctness and multi-object recognition performance.

\noindent\textbf{Thinking Format Reward.} This reward constrains the model to output a thinking process between the \texttt{<think>} and \texttt{</think>} tags, and the final answer between the \texttt{<answer>} and \texttt{</answer>} tags.

\noindent\textbf{Answer Format Reward.} We adopt bounding boxes $\{B_i\}_{i=1}^{N}$, positive points $\{P^{\text{central}}_i\}_{i=1}^{N}$, and negative points $\{P^{\text{neg}}_i\}_{i=1}^{N}$ as the answer format to improve training efficiency and spatial precision. So this reward restrict the model output answer in \texttt{[ \{ 'bbox\_2d': [x\_1, y\_1, x\_2, y\_2], 'point\_pos': [x\_1, y\_1, 1], 'point\_neg': [x\_2, y\_2, 0] \}, ... ]}.

\noindent\textbf{Non Repeat Reward.} To avoid repeated patterns, we split the reasoning process into individual sentences and prioritize those with unique or non-repetitive thinking processes.

\noindent\textbf{Bboxes IoU Reward.} Given a set of $N$ ground-truth bounding boxes and $K$ predicted bounding boxes, this reward computes their optimal one-to-one matched Intersection-over-Union (IoU) scores. For each IoU greater than 0.5, we increment the reward by \(\frac{1}{\max(N, K)}\).

\noindent\textbf{Bboxes L1 Reward.} Given a set of $N$ ground-truth bounding boxes and $K$ predicted bounding boxes, this reward computes the one-to-one matched L1 distances. For each L1 distance less than a threshold of 10 pixels, we increment the reward by
\(\frac{1}{\max(N, K)}\).

\noindent\textbf{Points L1 Reward.} Given a set of $N$ ground-truth points and $K$ predicted points, this reward computes their one-to-one matched L1 distances. For each L1 distance less than a threshold of 30 pixels, we increment the reward by
\(\frac{1}{\max(N, K)}\).

\section{Task Difficulty Scoring Details}
\label{sec:Task Difficulty Scoring Details}
\subsection{Prompts for Scoring}
\label{sec:Prompts for Scoring}
\begin{figure*}[t!]
    \centering
    \includegraphics[width=1\linewidth]{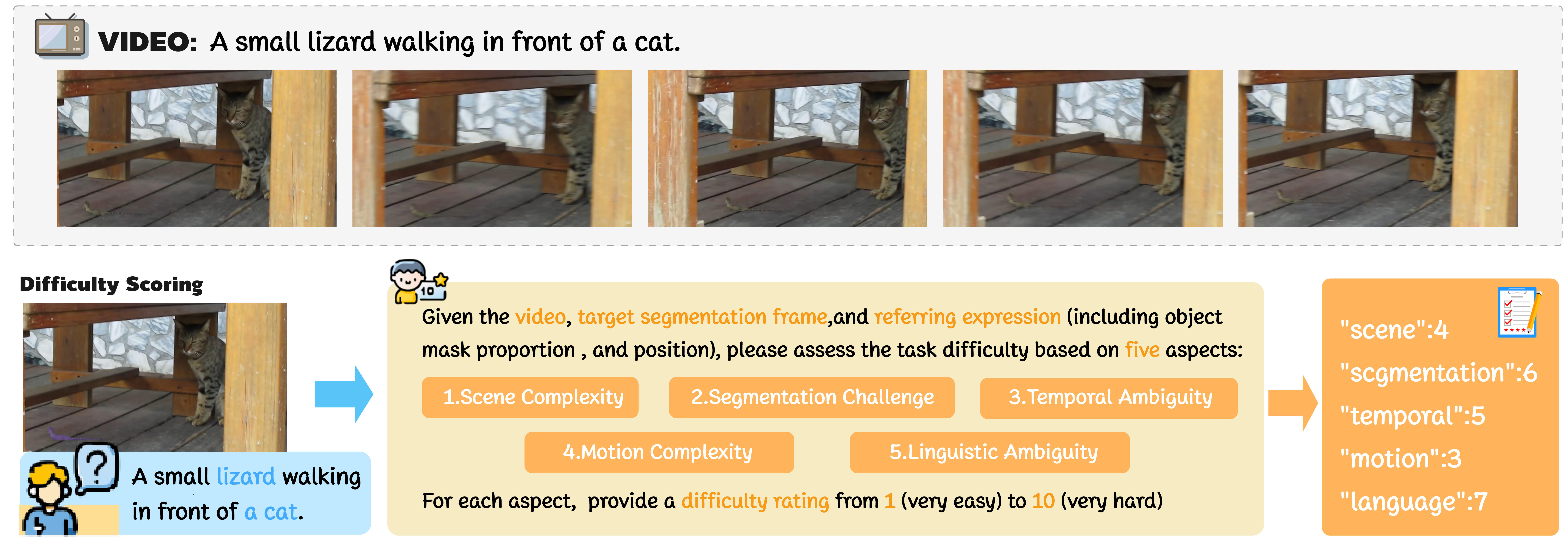}
    \caption{Design on the Difficulty Scoring scheme.}
    \label{fig:Design on the Difficulty Scoring scheme}
\end{figure*}

To support reinforcement learning in VideoSeg-R1, we assign a task difficulty score to each of the 15,000 training samples, which are manually selected from the Ref-YouTube-VOS~\cite{seo2020urvos}, MeViS~\cite{dingMeViSLargescaleBenchmark2023}, and Ref-DAVIS17~\cite{pont-tuset2017DAVISChallenge2018} datasets. Among them, approximately 60\% are from MeViS, 30\% from Ref-YouTube-VOS, and 10\% from Ref-DAVIS17. To achieve this, we generate \textit{visual descriptions} based on the segmentation mask properties (e.g., size and position) and \textit{textual descriptions} derived from the referring expressions (e.g., expression length and the number of spatial terms). These descriptions are incorporated into the prompt. We then instruct Qwen2.5VL-72B~\cite{baiQwen25VLTechnicalReport2025} to rate each sample from five perspectives: (1) scene complexity, (2) segmentation challenge, (3) temporal ambiguity, (4) motion complexity, and (5) linguistic ambiguity. The final difficulty score is computed as the average of these five ratings. The prompt used for this scoring process is provided below. Details can be found in \cref{fig:Design on the Difficulty Scoring scheme}.

\begin{tcolorbox}[
    title=Difficulty Scoring Prompt,
    coltitle=white,
    colback=white,
    colframe=boxcolor,
    colbacktitle=boxcolor,
    fonttitle=\bfseries,
    sharp corners,
    boxrule=2pt
]
You are an expert in reasoning segmentation evaluation.\\
Given the full video, a target segmentation frame, and the referring expression:  
\textcolor{blue}{\texttt{\{``Question''\}}}, please assess the task difficulty based on the following five aspects:

  \texttt{1.Scene Complexity:} \\
 - How many objects are visible in the segmentation frame? \\
    - How many of them are potentially related or visually similar to the target?

\texttt{2.Segmentation Challenge:} \\
    - What is the size and position of the target object in the segmentation frame? \\
    - Are there occlusions, overlaps, or visually similar objects nearby? \\
    - Is the mask describing the whole object or just a part? \\
    \textcolor{blue}{\texttt{\{``Visual Description''\}}}

 \texttt{3.Temporal Ambiguity:} \\
    - Does the target only appear at a specific moment or short segment of the video, requiring precise temporal localization? \\
    - Does the referring expression involve reasoning over the order of events to identify the target? \\
    - Is it necessary to understand causal or sequential relationships in the video timeline to resolve the reference?

 \texttt{4.Motion Complexity:} \\
    - Is the target moving quickly, deforming, or interacting with other objects? \\
     - Does the target frequently leave and re-enter the frame, or show complex movement patterns?

\texttt{5.Linguistic Ambiguity:} \\
    - Does the referring expression explicitly and clearly identify the target object? \\
    - Or does it require commonsense, multi-step reasoning, or disambiguation between multiple similar entities? \\
    \textcolor{blue}{\texttt{\{``Textual Description''\}}}

For each aspect, please provide a difficulty rating from 1 (very easy) to 10 (very hard), and summarize in the following Python dictionary format.\\
\textit{e.g.,} \texttt{\{"scene": 4, "segmentation": 6, "temporal": 5, "motion": 3, "language": 7\}}

\end{tcolorbox}
\subsection{Lable Statistics}
We divide the task difficulty score \( D \in [1, 10] \) into three levels using two thresholds: \(\tau_{\text{easy}} = 3.0\) and \(\tau_{\text{hard}} = 6.0\). Specifically, samples with \( D \leq \tau_{\text{easy}} \) are categorized as \textbf{easy}, those with \(\tau_{\text{easy}} < D \leq \tau_{\text{hard}} \) as \textbf{medium}, and those with \( D > \tau_{\text{hard}} \) as \textbf{hard}. Based on this criterion, we partition the 15{,}000 manually selected training samples into three difficulty levels. The distribution of samples across the three levels is shown in Table~\ref{tab:difficulty_distribution}.

\begin{table}[ht]
\centering

\begin{tabular}{lccc}
\toprule
\textbf{Dataset} & \textbf{Easy} & \textbf{Medium} & \textbf{Hard} \\
\midrule
MeViS (60\%) & 2600 & 5200 & 1200 \\
Ref-YouTube-VOS (30\%) & 1300 & 2700 & 500 \\
Ref-DAVIS17 (10\%) & 400 & 1000 & 100 \\
\midrule
\textbf{Total (15,000)} & \textbf{4300} & \textbf{8900} & \textbf{1800} \\
\bottomrule
\end{tabular}

\caption{Difficulty distribution of the 15,000 manually selected training samples across datasets.}
\label{tab:difficulty_distribution}
\end{table}
\section{Budget Mapping}
\label{sec:budget mapping}
The reasoning token budget \( L_{\text{budget}} \) is assigned based on the difficulty level as follows:
\[
L_{\text{budget}} =
\begin{cases}
L_{\text{easy}} = 96, & \text{if } D \leq \tau_{\text{easy}} \quad \text{(Easy)} \\
L_{\text{mid}} = 176, & \text{if } \tau_{\text{easy}} < D \leq \tau_{\text{hard}} \quad \text{(Medium)} \\
L_{\text{hard}} = 256, & \text{if } D > \tau_{\text{hard}} \quad \text{(Hard)}
\end{cases}
\]

\section{Implementation Details}
\label{sec:Implementation Details}
During reinforcement learning fine-tuning, we adopt the GRPO~\cite{GRPO} algorithm for optimization, with the KL divergence coefficient set to $1 \times 10^{-3}$ and the soft length penalty factor $\beta$ set to $2 \times 10^{-3}$ to enable adaptive reasoning length control based on task difficulty. To enhance cross-domain robustness, all input images are resized to $840 \times 840$ resolution before being fed into the multi-modal large language model during both training and evaluation stages. During training, we disable the Hierarchical Text-guided Frame Sampler (HTFS) and instead randomly sample one target frame, 8–12 reference frames, and a key segment of corresponding length from each sample to ensure broader training coverage. In the inference stage, HTFS is activated and combined with an adaptive global sampling strategy, where 12 reference frames are selected to provide global context. The relative segment length threshold $\delta$ is set to 0.3 to constrain the temporal span of key segments. The full training process takes approximately 1.5 days.
\section{Qualitative Results}
\label{sec:Qualitative Results}
\begin{figure*}[t!]
    \centering
    \includegraphics[width=1\linewidth]{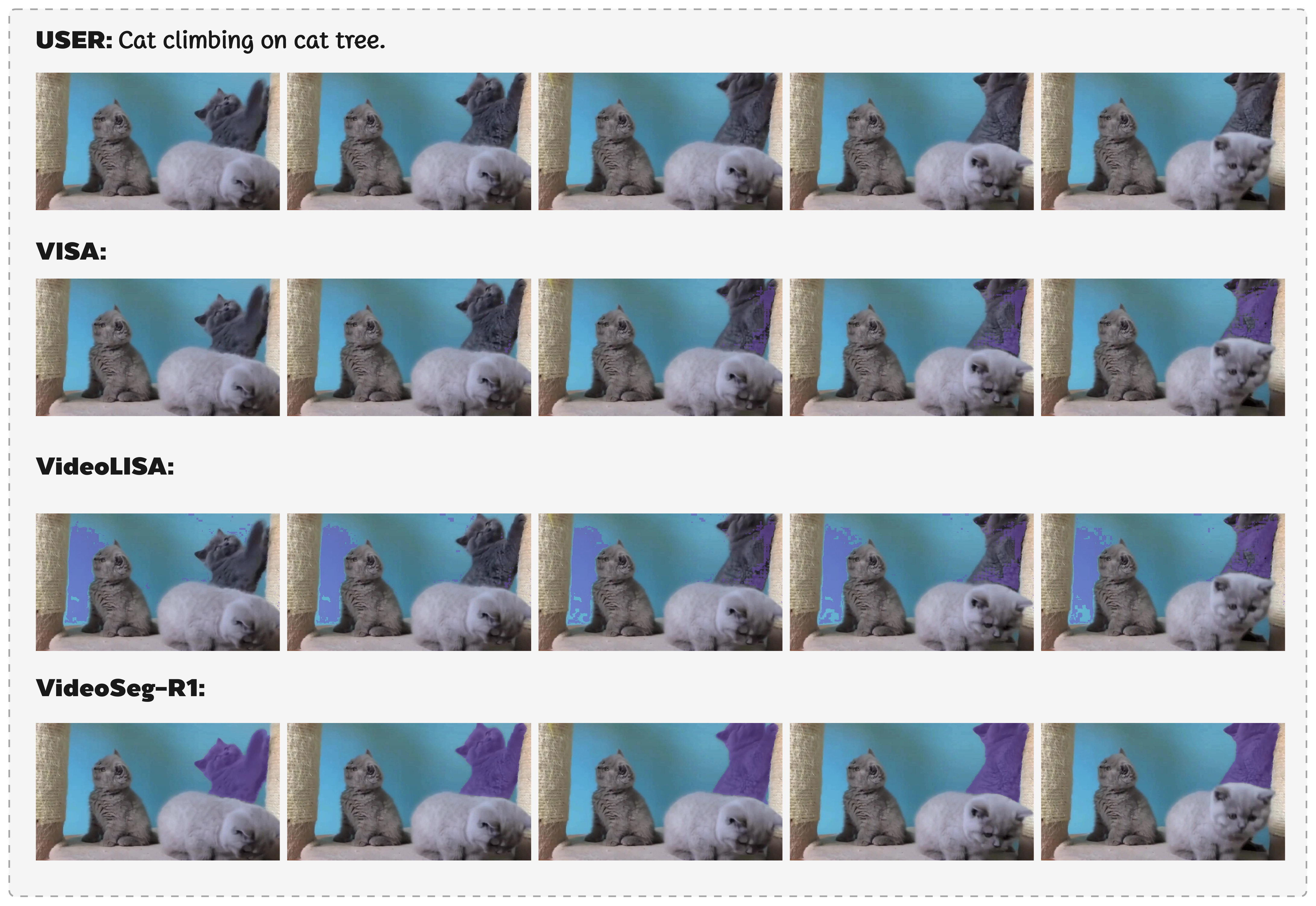}
    \caption{Qualitative comparisons between VideoSeg-R1, VideoLISA, and VISA.}
    \label{fig:Qualitative comparison}
\end{figure*}

\begin{figure*}[t!]
    \centering
    \includegraphics[width=0.9\linewidth]{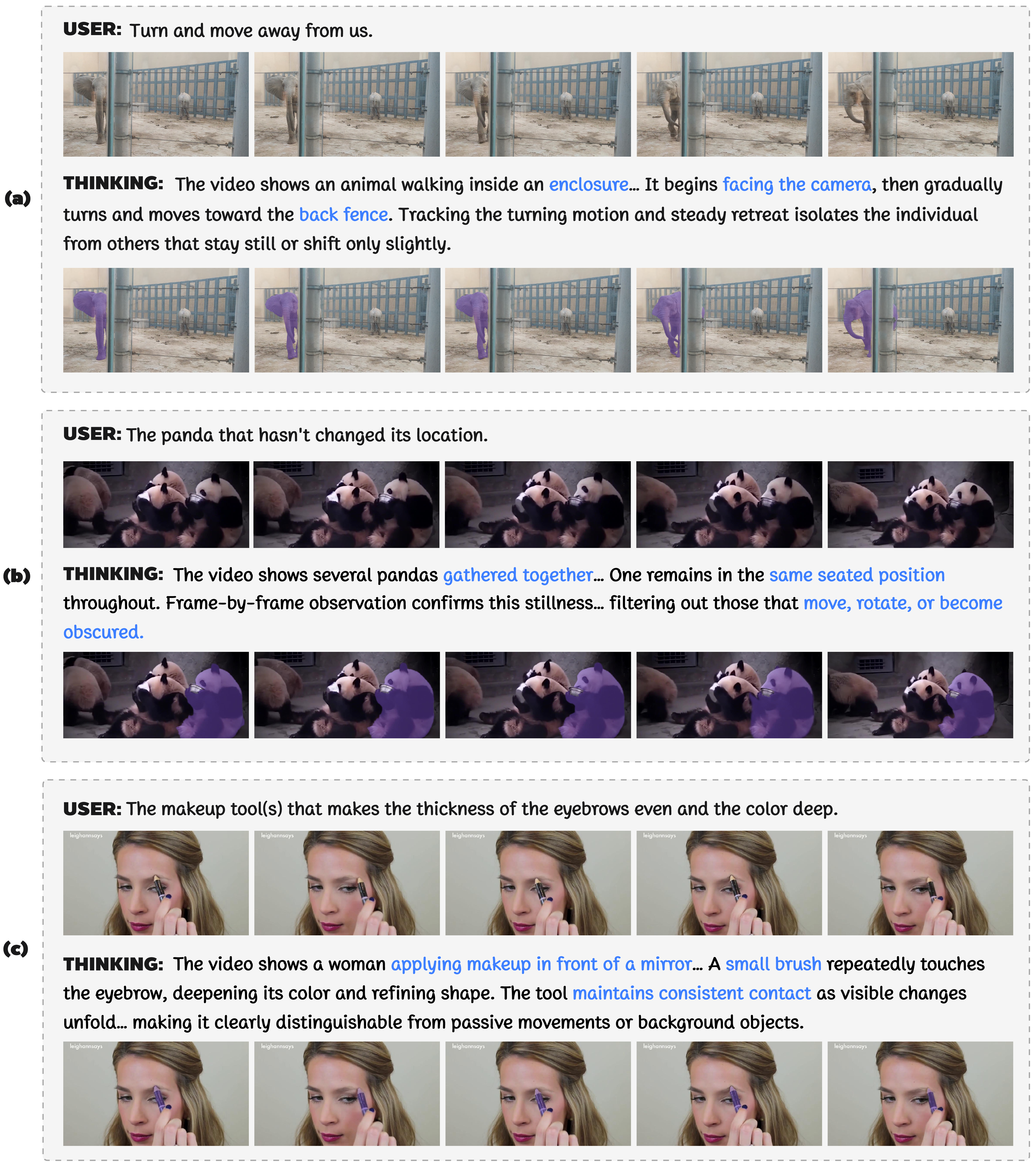}
    \caption{More qualitative examples of VideoSeg-R1.}
    \label{fig:More qualitative examples}
\end{figure*}

In \cref{fig:Qualitative comparison}, we present a visual comparison among VISA, VideoLISA, and our proposed VideoSeg-R1. When handling the complex scenario of ``cat climbing on cat tree", VISA shows inconsistent segmentation, particularly failing to track the cat's contour accurately during motion. VideoLISA improves temporal consistency to some extent but still struggles with occluded regions in certain frames. In contrast, our VideoSeg-R1 maintains clear and stable segmentation boundaries throughout the entire sequence, accurately capturing the cat's movement on the tree, demonstrating stronger temporal modeling ability and robustness. In \cref{fig:More qualitative examples}, we provide more extensive qualitative examples of VideoSeg-R1.
\section{Ablation Studies}
\label{sec:Ablation Studies}
\begin{table*}[t]
\centering

\setlength{\tabcolsep}{3pt}
\begin{tabular}{l
                |cccc
                |cccc
                |ccc
                |ccc}
\toprule
\multirow{2}{*}{\textbf{Mode}} 
& \multicolumn{4}{c|}{\textbf{refCOCO}}
& \multicolumn{4}{c|}{\textbf{refCOCO+}}
& \multicolumn{3}{c|}{\textbf{refCOCOg}}
& \multicolumn{3}{c}{\textbf{ReasonSeg}} \\
\cmidrule(lr){2-5} \cmidrule(lr){6-9} \cmidrule(lr){10-12} \cmidrule(lr){13-15}
& Token$\downarrow$ & val$\uparrow$ & testA$\uparrow$ & testB$\uparrow$
& Token$\downarrow$ & val$\uparrow$ & testA$\uparrow$ & testB$\uparrow$
& Token$\downarrow$ & val(U)$\uparrow$ & test(U)$\uparrow$
& Token$\downarrow$ & gIoU$\uparrow$ & cIoU$\uparrow$ \\
\midrule
\textbf{w/o} 
& 63.1 & 76.5 & 79.6 & 73.0
& 70.3 & 68.3 & 73.8 & 60.8
& 79.2 & 70.4 & 71.0
& 82.3 & 65.1 & 63.5 \\

\textbf{w/}  
& 37.2 & 78.2 & 82.3 & 75.1
& 40.5 & 71.8 & 76.1 & 64.7
& 42.8 & 73.1 & 74.1
& 44.2 & 69.8 & 68.2 \\
\bottomrule
\end{tabular}

\caption{Ablation study of Soft Length Penalty on \textbf{image datasets}.}
\label{tab:ablation-image-token-fix}
\end{table*}

\begin{table*}[ht]
\centering

\begin{tabular}{l|ccc|c|c|c}
\hline
\textbf{Method} & \multicolumn{3}{c|}{\textbf{Training Data}} & \textbf{Image} & \textbf{Video} & \textbf{Avg.} \\
 & Ref-YouTube-VOS & Ref-DAVIS17 & MeViS & ReasonSeg & ReVOS & \\
\hline
VideoSeg-R1  & \cmark &        &        & 63.8 & 54.9 &59.4 \\
             & \cmark & \cmark &        & 65.2 & 57.3 & 61.3 \\
             & \cmark & \cmark & \cmark & \textbf{69.8} & \textbf{61.1} &\textbf{65.5}\\
\hline
\end{tabular}
\caption{Performance comparison on different training data.}
\label{tab:training_data_ablation}
\end{table*}

\noindent\textbf{Ablation on Token Budget Allocation.}  
We further conduct ablation studies on different token budget configurations used for the soft length penalty.  
Specifically, we define three discrete budget levels: \( L_{\text{easy}} = 96 \), \( L_{\text{mid}} = 176 \), and \( L_{\text{hard}} = 256 \), and evaluate the performance under various parameter combinations.  
As shown in Table~\ref{tab:token-budget-updated}, our method achieves significant reductions in token usage while maintaining or improving segmentation quality.  
In particular, the configuration \((96,176,256)\) yields the best performance on both ReVOS and ReasonSeg, along with the lowest token consumption, demonstrating the robustness of our proposed budget design.
\begin{table}[htbp]
\centering
\begin{tabular}{c|cc|cc}
\hline
\multirow{2}{*}{\textbf{Budget}} 
& \multicolumn{2}{c|}{\textbf{ReVOS}} 
& \multicolumn{2}{c}{\textbf{ReasonSeg}} \\
\cline{2-5}
 & \textbf{Token$\downarrow$} & \textbf{J\&F$\uparrow$} 
 & \textbf{Token$\downarrow$} & \textbf{gIoU$\uparrow$} \\
\hline
w/o & 89.9 & 57.3 & 82.3 & 65.1 \\
(64,96,176) & 72.3 & 60.6 & 69.2 & 67.5 \\
(64,176,256) & 74.7 & 60.3 & 64.7 & 68.0 \\
(96,176,256) & \textbf{56.1} & \textbf{61.1} &\textbf{44.2} & \textbf{69.8} \\
(96, 192, 256) & 67.2 & 58.5 & 70.3 & 66.3 \\
\hline
\end{tabular}
\caption{Ablation on the token budget allocation details.}
\label{tab:token-budget-updated}
\end{table}

\noindent\textbf{Ablation on Difficulty Splits.}  
We also investigate the impact of difficulty granularity by varying the number of difficulty levels used during training.  
In the 2-level setting, medium and hard samples are merged and assigned a unified token budget.  
In contrast, the 4-level setting further subdivides the medium category and applies more fine-grained length constraints.  
As shown in Table~\ref{tab:difficulty-splits-final}, the 3-level split offers the best trade-off between segmentation accuracy and token efficiency.  
This configuration not only achieves the lowest token cost but also obtains the highest evaluation metrics, verifying the effectiveness of our difficulty-aware design strategy.
\begin{table}[htbp]
\centering
\begin{tabular}{c|cc|cc}
\hline
\multirow{2}{*}{\textbf{Difficulty}} 
& \multicolumn{2}{c|}{\textbf{ReVOS}} 
& \multicolumn{2}{c}{\textbf{ReasonSeg}} \\
\cline{2-5}
 & \textbf{Token$\downarrow$} & \textbf{J\&F$\uparrow$} 
 & \textbf{Token$\downarrow$} & \textbf{gIoU$\uparrow$} \\
\hline
w/o & 89.9 & 57.3 & 82.3 & 65.1 \\
2 & 61.7 & 59.3 & 61.5 & 67.1 \\
3 & \textbf{56.1} & \textbf{61.1} &\textbf{44.2} & \textbf{69.8} \\
4 & 73.8 & 60.5 & 76.8 & 68.8 \\
\hline
\end{tabular}
\caption{Ablation of the difficulty splits.}
\label{tab:difficulty-splits-final}
\end{table}

\noindent\textbf{Ablation on Soft Length Penalty on image datasets.}
As shown in Table~\ref{tab:ablation-image-token-fix}, applying the soft length penalty consistently improves both efficiency and accuracy across all image datasets. On refCOCO, token usage drops significantly from 63.1 to 37.2, accompanied by consistent gains in all splits. Similar improvements are observed on refCOCO+ and refCOCOg, demonstrating the general effectiveness of the penalty. On ReasonSeg, the gIoU increases from 65.1 to 69.8, further confirming that the soft length penalty enhances segmentation quality while reducing unnecessary reasoning tokens.

\noindent\textbf{Ablation on Training Datasets.}
Table~\ref{tab:training_data_ablation} presents the impact of different training data combinations on the performance of VideoSeg-R1. When trained solely on Ref-YouTube-VOS and Ref-DAVIS17, the model achieves an average performance of 61.3. By further incorporating the MeViS dataset, the performance improves significantly, reaching an average score of 65.5. This demonstrates that diverse video data contributes positively to enhancing the model's generalization ability and reasoning performance.
\section{Failure Cases}
\label{sec:Failure Cases}
\begin{figure*}[t!]
    \centering
    \includegraphics[width=0.9\linewidth]{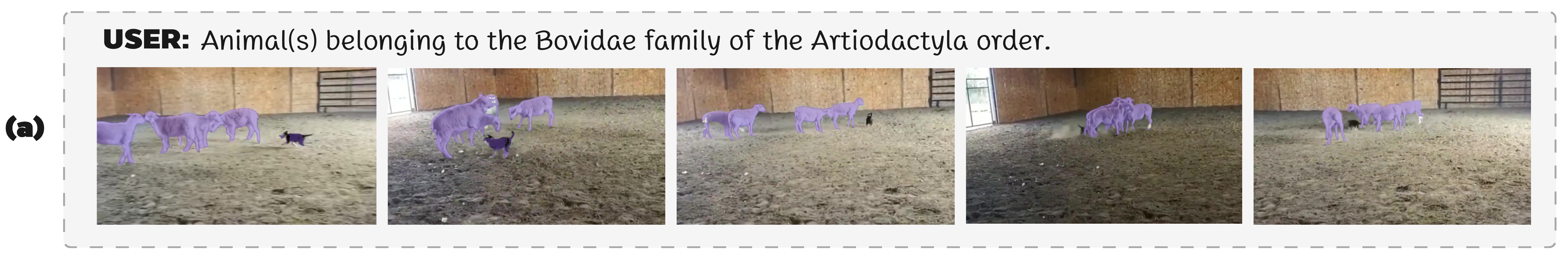}
    \caption{Failure case of VideoSeg-R1.}
    \label{fig:Failure cases of VideoSeg-R1}
\end{figure*}

Although our VideoSeg-R1 demonstrates strong performance in video reasoning segmentation tasks, there is still room for improvement. As illustrated in ~\cref{fig:Failure cases of VideoSeg-R1}, the model incorrectly segments a dog in the scene and misclassifies it as an \textit{``animal(s) belonging to the Bovidae family of the Artiodactyla order''}. We hypothesize that this error primarily stems from the model's limited understanding of complex world knowledge, particularly when it comes to semantic distinctions in animal morphology. This issue becomes more prominent in long-shot frames, where reduced resolution and visual detail make it easier for the model to confuse species with similar body size, limb structure, or silhouette. This case highlights that VideoSeg-R1 still faces challenges in scenarios requiring background knowledge reasoning, and further improvements are needed to enhance its fine-grained discriminative capabilities.



\end{document}